\documentclass[conference]{IEEEtran}
\IEEEoverridecommandlockouts
\usepackage{cite}
\usepackage{amsmath,amssymb,amsfonts}
\usepackage{algorithmic}
\usepackage{graphicx}
\usepackage{textcomp}
\usepackage{xcolor}
\usepackage{booktabs}
\def\BibTeX{{\rm B\kern-.05em{\sc i\kern-.025em b}\kern-.08em
    T\kern-.1667em\lower.7ex\hbox{E}\kern-.125emX}}
\makeatletter 
\newcommand{\linebreakand}{%
  \end{@IEEEauthorhalign}
  \hfill\mbox{}\par
  \mbox{}\hfill\begin{@IEEEauthorhalign}
}
\makeatother 

\begin{document}

\title{SemHash-LLM: A Multi-Granularity Semantic Hashing Framework for Document Deduplication\\}

\author{
\IEEEauthorblockN{Xinyi Fang *}
\IEEEauthorblockA{\textit{Independent Researcher} \\
Shanghai, China \\
fangxy.carol@gmail.com}
\and
\IEEEauthorblockN{Kejian Tong}
\IEEEauthorblockA{\textit{Independent Researcher} \\
Mukilteo, USA \\
tongcs2021@gmail.com}
\and
\IEEEauthorblockN{Jiabei Liu}
\IEEEauthorblockA{\textit{Northeastern University} \\
Oakland, USA \\
liu.jiabe@northeastern.ed}
\and
\linebreakand
\IEEEauthorblockN{Tao Ning}
\IEEEauthorblockA{\textit{Syracuse University} \\
San Jose, USA \\
ntgd1102@gmail.com}
\and
\IEEEauthorblockN{Yuhang He}
\IEEEauthorblockA{\textit{Independent Researcher} \\
Chicago, USA \\
yuhang.he@outlook.com}
}

\maketitle

\begin{abstract}
Large scale document deduplication must preserve semantic equivalence while remaining efficient over massive corpora. We present SemHash LLM, a multi granularity framework that unifies semantic projection hashing, attention weighted MinHash, contrastive boundary learning, and selective LLM based adjudication. The method combines character, token, and document level signals through gated fusion, then applies a cascaded filtering pipeline for efficient candidate reduction. Semantic projection hashing learns compact binary codes in distilled LLM embedding space, while attention weighted MinHash suppresses boilerplate and emphasizes informative content. Adaptive decision boundaries and uncertainty estimation further improve robustness across template pollution, short text perturbation, containment, and viral fragments. Experiments show that SemHash LLM achieves strong duplicate detection quality with less than one percent neural verification cost.
\end{abstract}


\section{Introduction}
Modern data pipelines face a persistent dilemma, large web corpora demand aggressive deduplication for efficiency and data quality, yet strict lexical filtering often removes useful semantic variants while semantic methods are usually too expensive at scale. This tension has become more important as recent language model studies show that data redundancy strongly affects memorization, optimization, and downstream quality \cite{lee2022deduplicating}.Similar concerns about resilience under dynamic operating conditions have also been studied in logistics and routing systems \cite{xue2026resilient}. At the same time, open pretraining corpora continue to grow in size and heterogeneity, making scalable curation a first order systems problem \cite{gao2020pile}.

Current solutions struggle because they optimize only one side of the trade off. Exact and near exact fingerprinting methods are fast, but they are brittle under paraphrase, template wrapping, and small character perturbations. Pure embedding based retrieval is more semantically aware, but its computational and thresholding costs are difficult to control across billions of documents.Related efforts on efficient large model inference also explore multi-stage compression and acceleration strategies that balance accuracy with deployment cost, including adaptive quantization, token pruning, and decode-time optimization \cite{zhouunified}. These challenges are amplified in modern corpora that mix long documents, short snippets, boilerplate heavy pages, and containment relationships within the same pipeline \cite{soldaini2024dolma}.

We address this problem with SemHash LLM, a unified framework that combines learned semantic hashing, attention weighted lexical sketching, adaptive contrastive decision boundaries, and selective LLM based refinement. Our design breaks the conventional efficiency versus accuracy compromise by using a cascaded pipeline for pruning, multi granularity fusion for robustness, and uncertainty aware routing so that expensive reasoning is applied only to rare borderline cases.

\section{Related Work}
Recent work on retrieval and matching has shown that contextualized token interaction can improve fine grained similarity estimation beyond bag of words fingerprints. COIL demonstrated that exact lexical matching can be strengthened with contextualized token representations, while ColBERTv2 improved efficiency through lightweight late interaction for large scale retrieval \cite{gao2021coil} \cite{santhanam2022colbertv2}. These studies motivate deduplication systems that preserve lexical precision while incorporating richer semantic structure.

Efficient semantic modeling has also advanced through compact encoders and contrastive representation learning.SimCSE further established that simple contrastive objectives can produce strong sentence embeddings with clear similarity structure \cite{gao2021simcse}. Our method builds on these ideas by learning binary semantic projections and adaptive duplicate boundaries rather than relying on fixed embedding thresholds alone.

A third line of work concerns model based judgment and confidence aware evaluation. MT Bench and Chatbot Arena provided evidence that strong language models can act as practical judges for nuanced comparisons \cite{zheng2023judging}. G Eval showed that carefully designed prompts can turn LLMs into reliable structured evaluators \cite{liu2023g}. SelfCheckGPT highlighted the value of uncertainty signals when trusting generative model outputs \cite{manakul2023selfcheckgpt}. We extend this direction to document deduplication by using LLM judgment only for uncertain candidate pairs identified by the automated pipeline.Related multi-agent LLM systems also show the value of hierarchical specialization and evidence-grounded reasoning in complex troubleshooting workflows \cite{yan2026prism}.Related work on dynamic retrieval and selective tool invocation also shows that LLM-based systems can route between retrieved evidence and external APIs when evidence is insufficient, as demonstrated by DynaRAG \cite{liang2026dynarag}.

\section{Methodology}
\begin{figure}[htbp]
\centering
\includegraphics[width=0.5\textwidth]{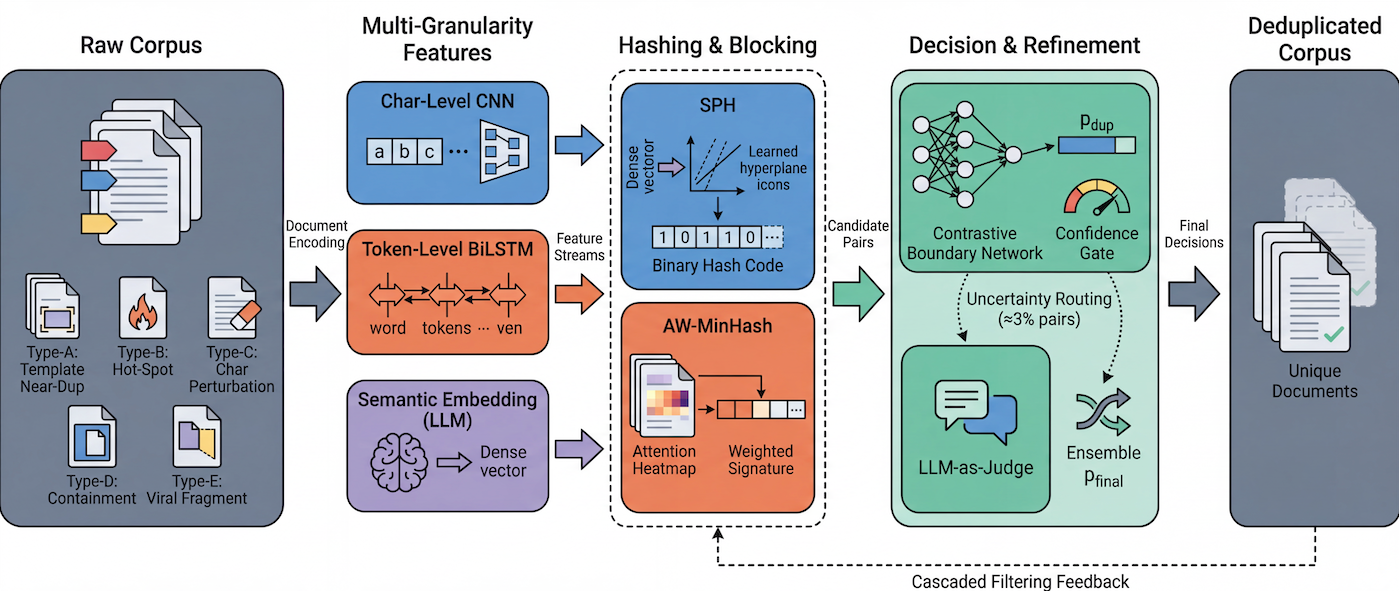}
\caption{Overview of the SemHash-LLM framework. Documents are encoded into multi-granularity features (character, token, semantic levels), processed through Semantic Projection Hashing and Attention-Weighted MinHash for efficient candidate generation, and refined via Contrastive Boundary Learning with LLM-as-Judge for borderline cases. The pipeline handles five deduplication types (A--E) within a unified architecture.}
\label{fig:149_1}
\end{figure}

In this section, we present SemHash-LLM, a semantic-aware hierarchical deduplication framework that bridges lexical fingerprinting with deep semantic understanding through LLM integration. As illustrated in Fig.~\ref{fig:149_1}, our framework integrates five core components operating across multiple representational granularities—from character perturbations to token patterns to document-level semantics—with adaptive weighting determined by the redundancy type encountered. Semantic Projection Hashing learns locality-sensitive hash functions in the LLM embedding space, balancing similarity preservation with quantization loss and orthogonality constraints. Attention-Weighted MinHash addresses template pollution by extracting importance weights from transformer attention patterns into the consistent weighted sampling framework. Contrastive Boundary Learning provides self-supervised adaptive decision boundaries through margin-based optimization over document pair distributions. For borderline cases with high uncertainty, an LLM-as-Judge mechanism fuses structured model judgments with automated predictions via confidence-weighted ensemble. These components are unified within a Multi-Granularity Fusion Network employing gated attention to aggregate character-level, token-level, and semantic-level features.These components are unified within a Multi-Granularity Fusion Network employing gated attention to aggregate character-level, token-level, and semantic-level features \cite{xu2026pyramid}. Trillion-scale efficiency is achieved through a cascaded filtering pipeline combining Bloom filters, semantic hash blocking, and attention-weighted LSH. The framework handles five deduplication categories: template-wrapped near-duplicates, hot-spot documents with distribution skew, short texts with character perturbations, parent-child containment hierarchies, and high-frequency viral fragments.

\subsection{Semantic Projection Hashing}

\begin{figure}[htbp]
\centering
\includegraphics[width=0.5\textwidth]{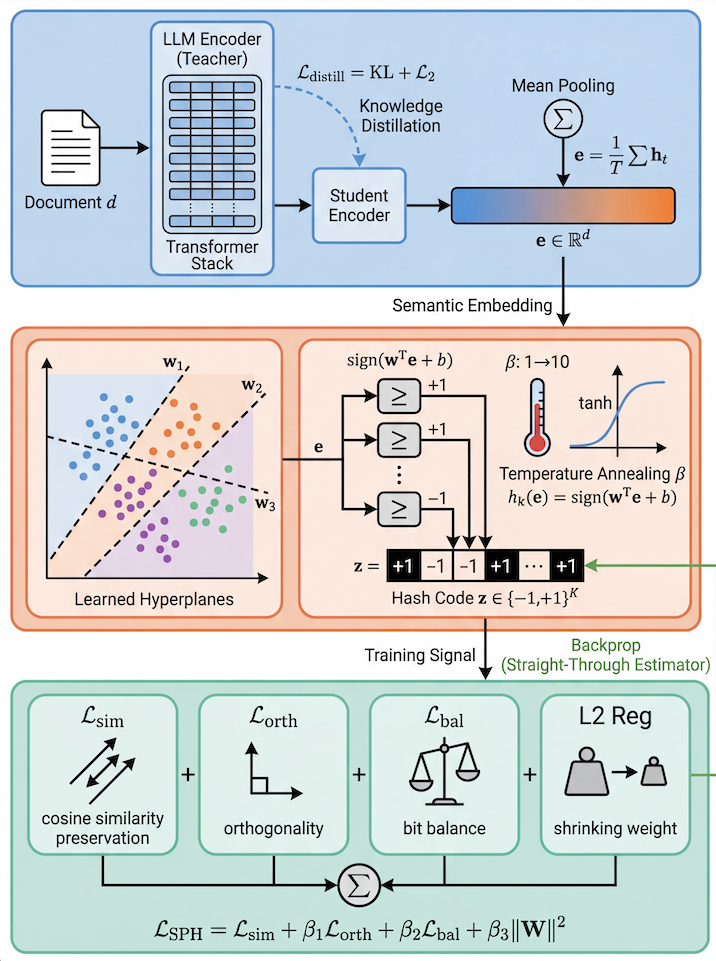}
\caption{Semantic Projection Hashing. A student encoder distilled from the LLM teacher produces semantic embeddings, which are mapped to compact binary codes via learned hyperplane partitions. The training objective jointly optimizes similarity preservation, orthogonality, and bit balance.}
\label{fig:149_2}
\end{figure}

Traditional locality-sensitive hashing methods such as MinHash and SimHash operate on discrete token or n-gram representations, fundamentally limiting their ability to capture semantic relationships. A document about ``automobile manufacturing'' and another about ``car production'' would produce entirely different hash signatures despite conveying nearly identical meaning. This limitation motivated our development of Semantic Projection Hashing, which learns hash functions that operate directly in the continuous LLM embedding space. The complete SPH training pipeline is depicted in Fig.~\ref{fig:149_2}.

\subsubsection{Embedding Extraction and Efficiency Considerations}

Given a document $d$ with content $c$, we first obtain a semantic representation through a pre-trained LLM encoder. The straightforward approach of using the final hidden state of a special classification token proved suboptimal in our experiments, as it tends to capture global document statistics rather than fine-grained semantic content. Instead, we employ mean pooling over all token representations:

\begin{equation}
\mathbf{e} = \frac{1}{T} \sum_{t=1}^{T} \mathbf{h}_t^{(L)}
\label{eq:embedding}
\end{equation}

\noindent where $\mathbf{h}_t^{(L)} \in \mathbb{R}^{d}$ is the hidden state at the final layer $L$ for token $t$, and $T$ is the sequence length. However, applying a full-scale LLM encoder to billions of documents would be computationally intractable.This efficiency bottleneck is consistent with roofline-style analyses of memory-bound LLM inference on Arm CPUs \cite{zhouroofline}. We address this through knowledge distillation, training a lightweight student encoder to mimic the semantic representations of a larger teacher model:

\begin{equation}
\mathcal{L}_{\text{distill}} = \text{KL}\left(p_{\text{tea}}(\mathbf{e}) \| p_{\text{stu}}(\mathbf{e})\right) + \alpha \|\mathbf{e}_{\text{tea}} - \mathbf{e}_{\text{stu}}\|_2^2
\label{eq:distill}
\end{equation}

The combination of KL divergence and L2 distance proved crucial for maintaining both distributional consistency and point-wise accuracy. Using either term alone resulted in degraded performance on downstream deduplication tasks, with KL-only training producing embeddings that preserved ranking but lost magnitude information important for threshold-based decisions.

\subsubsection{Learnable Hash Function Design}

With semantic embeddings in hand, we need to convert them into compact binary codes suitable for efficient similarity search. We learn a family of $K$ hash functions, each implemented as a hyperplane partition of the embedding space:

\begin{equation}
h_k(\mathbf{e}) = \text{sign}\left(\mathbf{w}_k^{\top} \mathbf{e} + b_k\right)
\label{eq:hash}
\end{equation}

\noindent where $\mathbf{w}_k \in \mathbb{R}^{d}$ and $b_k \in \mathbb{R}$ are learnable parameters. The complete hash code concatenates all $K$ bits:

\begin{equation}
\mathbf{z} = [h_1(\mathbf{e}), h_2(\mathbf{e}), \ldots, h_K(\mathbf{e})] \in \{-1, +1\}^K
\label{eq:code}
\end{equation}

A critical design decision involves the training objective for these hash functions. Random hyperplanes, as used in traditional SimHash, provide theoretical guarantees but ignore the specific structure of the document embedding distribution. We instead optimize the hash functions to satisfy a semantic preservation property:

\begin{equation}
\Pr\left[h_k(\mathbf{e}_i) = h_k(\mathbf{e}_j)\right] \approx \frac{1 + \cos(\mathbf{e}_i, \mathbf{e}_j)}{2}
\label{eq:property}
\end{equation}

This is achieved through a similarity preservation loss that encourages similar documents to produce similar hash codes:

\begin{equation}
\mathcal{L}_{\text{sim}} = \sum_{i,j} \left(S_{ij} - \frac{1}{K}\mathbf{z}_i^{\top}\mathbf{z}_j\right)^2
\label{eq:sim_loss}
\end{equation}

\noindent where $S_{ij} = \cos(\mathbf{e}_i, \mathbf{e}_j)$ is the target semantic similarity. However, directly optimizing this objective encounters the gradient discontinuity problem due to the sign function. We address this through the straight-through estimator during training, using $\tanh$ with temperature annealing as a smooth approximation:

\begin{equation}
\tilde{h}_k(\mathbf{e}) = \tanh\left(\beta \cdot (\mathbf{w}_k^{\top} \mathbf{e} + b_k)\right)
\label{eq:smooth}
\end{equation}

The temperature $\beta$ is gradually increased during training, starting from 1.0 and reaching 10.0, which smoothly transitions from soft assignments to hard binary decisions.

\subsubsection{Orthogonality and Information Maximization}

A subtle but important issue arises when learning multiple hash functions jointly: without explicit constraints, they tend to converge to similar hyperplanes, producing redundant bits that waste the limited code capacity. We enforce diversity through orthogonality regularization:

\begin{equation}
\mathcal{L}_{\text{orth}} = \left\|\mathbf{W}^{\top}\mathbf{W} - \mathbf{I}_K\right\|_F^2
\label{eq:orth}
\end{equation}

\noindent where $\mathbf{W} = [\mathbf{w}_1, \ldots, \mathbf{w}_K] \in \mathbb{R}^{d \times K}$ stacks all hyperplane normal vectors. Additionally, we observe that hash codes often suffer from bit imbalance, where certain bits are predominantly positive or negative across the corpus, reducing effective capacity. We add a balance regularization term:

\begin{equation}
\mathcal{L}_{\text{bal}} = \sum_{k=1}^{K} \left(\frac{1}{n}\sum_{i=1}^{n} \tilde{h}_k(\mathbf{e}_i)\right)^2
\label{eq:balance}
\end{equation}

The complete training objective for Semantic Projection Hashing combines these components:

\begin{equation}
\mathcal{L}_{\text{SPH}} = \mathcal{L}_{\text{sim}} + \beta_1 \mathcal{L}_{\text{orth}} + \beta_2 \mathcal{L}_{\text{bal}} + \beta_3 \|\mathbf{W}\|_F^2
\label{eq:sph_total}
\end{equation}

\noindent where the final term provides standard L2 regularization to prevent overfitting. Through grid search on a validation set, we found $\beta_1 = 0.1$, $\beta_2 = 0.01$, and $\beta_3 = 0.001$ to provide a good balance across diverse document types.

\subsection{Attention-Weighted MinHash}

\begin{figure}[htbp]
\centering
\includegraphics[width=0.5\textwidth]{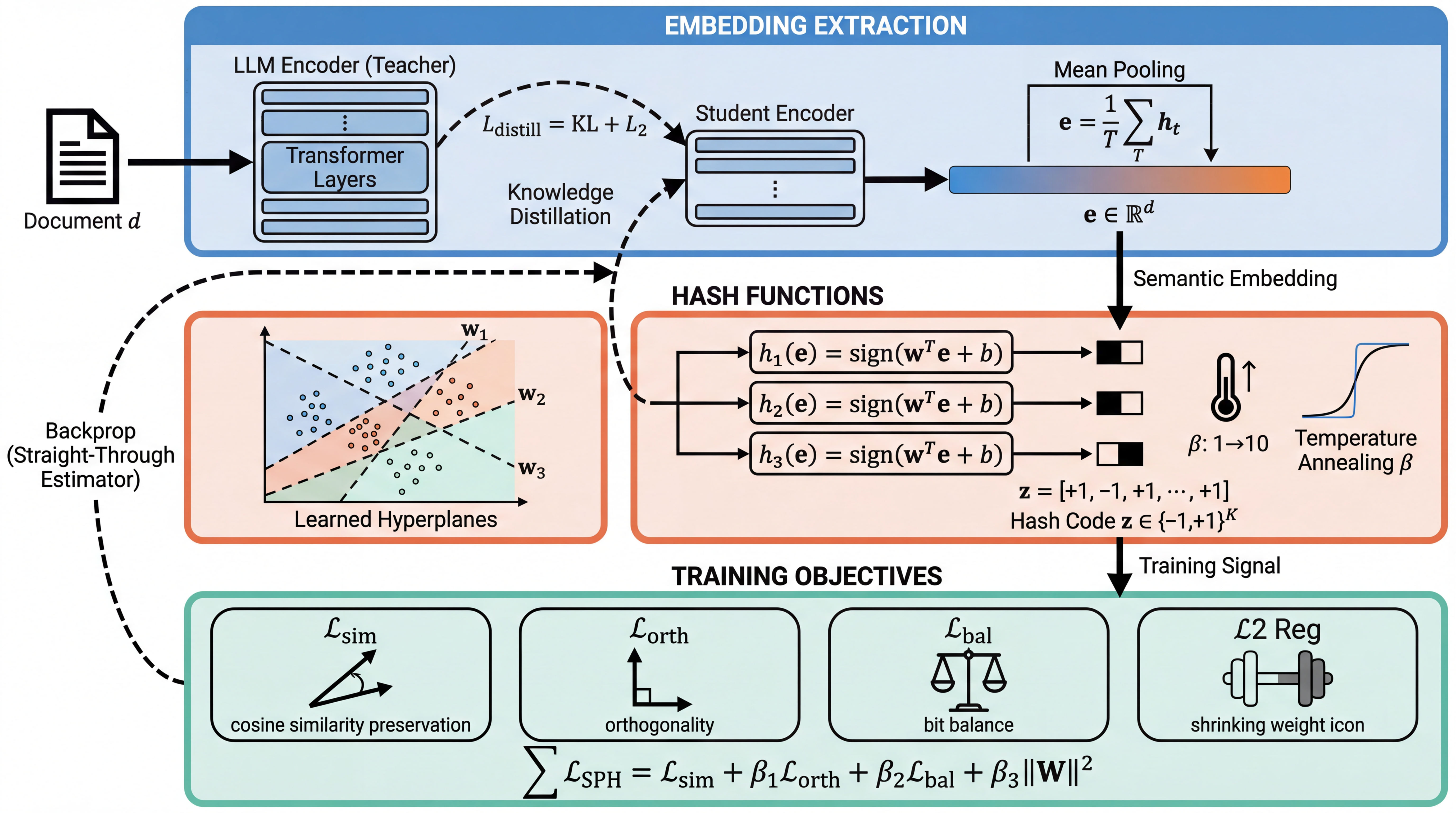}
\caption{Attention-Weighted MinHash pipeline. Aggregated multi-head attention scores highlight discriminative content while suppressing boilerplate. Combined with IDF weighting, the consistent weighted sampling generates importance-aware signatures, which are compared via adaptive LSH banding tuned by attention entropy.}
\label{fig:149_3}
\end{figure}

While Semantic Projection Hashing excels at capturing document-level semantic similarity, it may overlook fine-grained lexical patterns important for certain deduplication types. Standard MinHash addresses this through n-gram fingerprinting but treats all n-grams equally, which proves problematic when documents contain substantial boilerplate. Consider a news article wrapped in website navigation menus, cookie consent notices, and advertisement blocks: the discriminative news content may constitute only a small fraction of the total text, yet standard MinHash would give equal weight to boilerplate n-grams. Fig.~\ref{fig:149_3} illustrates how attention-derived importance weights.

\subsubsection{Extracting Attention as Importance Signal}

Our key insight is that transformer attention patterns naturally encode token importance: tokens receiving high attention from many other tokens are likely central to the document's meaning, while peripheral tokens like stopwords and boilerplate receive diffuse, unfocused attention. We extract multi-head attention scores from the LLM encoder:

\begin{equation}
\mathbf{A}^{(l,h)} = \text{softmax}\left(\frac{\mathbf{Q}^{(l,h)}{\mathbf{K}^{(l,h)}}^{\top}}{\sqrt{d_k}}\right)
\label{eq:attention}
\end{equation}

Rather than using attention from a single layer, which we found to be noisy and inconsistent, we aggregate across all layers and heads to obtain a robust importance estimate:

\begin{equation}
\bar{a}_t = \frac{1}{LH} \sum_{l=1}^{L} \sum_{h=1}^{H} \sum_{t'=1}^{T} A_{t',t}^{(l,h)}
\label{eq:agg_attn}
\end{equation}

This aggregated attention score indicates how much other positions attend to position $t$, serving as a proxy for content importance. An interesting finding during development was that earlier layers tend to capture syntactic patterns while later layers focus on semantic content. Uniform averaging across layers implicitly balances these complementary signals.

\subsubsection{Weighted MinHash Formulation}

For each character 5-gram $g$ spanning token positions $[i, j]$, we compute an importance weight that combines attention-derived importance with inverse document frequency:

\begin{equation}
w(g) = \left(\frac{1}{j-i+1} \sum_{t=i}^{j} \bar{a}_t\right) \cdot \log\frac{N + 1}{\text{df}(g) + 1}
\label{eq:gram_weight}
\end{equation}

The IDF term down-weights common n-grams that appear across many documents, complementing the attention signal which identifies within-document importance. We then apply consistent weighted sampling to generate MinHash signatures that respect these importance weights:

\begin{equation}
h_k^{\text{AW}}(d) = \arg\min_{g \in G_5(d)} \frac{-\log(U_{g,k})}{w(g) + \epsilon}
\label{eq:weighted_minhash}
\end{equation}

\noindent where $U_{g,k} \sim \text{Uniform}(0,1)$ is a random value determined by applying hash function $k$ to gram $g$, and $\epsilon = 10^{-8}$ prevents numerical instability. This formulation ensures that high-weight grams have proportionally higher probability of being selected as the minimum, effectively focusing the signature on discriminative content.

The resulting signatures satisfy a weighted Jaccard estimation property:

\begin{equation}
\Pr\left[h_k^{\text{AW}}(d_1) = h_k^{\text{AW}}(d_2)\right] = J_w(d_1, d_2)
\label{eq:wj_property}
\end{equation}

\noindent where the weighted Jaccard similarity is defined as:

\begin{equation}
J_w(d_1, d_2) = \frac{\sum_{g \in G_5(d_1) \cap G_5(d_2)} \min(w_1(g), w_2(g))}{\sum_{g \in G_5(d_1) \cup G_5(d_2)} \max(w_1(g), w_2(g))}
\label{eq:weighted_jaccard}
\end{equation}

\subsubsection{Adaptive LSH Configuration}

Standard LSH uses fixed band and row configurations across all documents, but we observed that documents with different attention entropy profiles require different sensitivity settings. Documents with concentrated attention on few key phrases exhibit clearer duplicate boundaries, allowing for more aggressive band configurations, while documents with diffuse attention over many equally important segments need finer discrimination.

We quantify this through attention entropy:

\begin{equation}
H_{\text{att}}(d) = -\sum_{t=1}^{T} \tilde{a}_t \log \tilde{a}_t
\label{eq:entropy}
\end{equation}

\noindent where $\tilde{a}_t = \bar{a}_t / \sum_{t'} \bar{a}_{t'}$ is the normalized attention distribution. The number of bands is then adapted:

\begin{equation}
b(d) = \text{clip}\left(b_0 + \lfloor \delta \cdot (H_{\text{att}}(d) - \bar{H}) \rfloor, b_{\min}, b_{\max}\right)
\label{eq:adaptive_band}
\end{equation}

\noindent where $b_0$ is the base band count, $\bar{H}$ is the corpus-wide mean entropy, and $\delta$ controls adaptation strength. The clipping ensures configurations remain within reasonable bounds to prevent extreme sensitivity variations.

\subsection{Contrastive Boundary Learning}

Fixed similarity thresholds (e.g., 0.8 for Jaccard, 0.9 for containment) assume uniform document characteristics. In practice, optimal thresholds vary significantly: technical documents tolerate higher lexical overlap, while generic web content requires stricter thresholds.

\subsubsection{Learning Adaptive Decision Boundaries}

We learn thresholds from data through a contrastive approach. A projection network maps document representations to a space where duplicates cluster together:

\begin{equation}
\mathcal{L}_{\text{con}} = -\sum_{(i,j) \in \mathcal{P}^+} \log \frac{\exp(\text{sim}(\mathbf{z}_i, \mathbf{z}_j)/\tau)}{\sum_{k \neq i} \exp(\text{sim}(\mathbf{z}_i, \mathbf{z}_k)/\tau)}
\label{eq:contrastive}
\end{equation}

\noindent where $\mathbf{z}_i = g_\theta([\mathbf{e}_i; \mathbf{f}_i^{\text{lex}}])$ combines semantic and lexical features, with temperature $\tau = 0.07$. A boundary network then predicts duplicate probability:

\begin{equation}
p_{\text{dup}}(d_i, d_j) = \sigma\left(f_\psi\left(\mathbf{z}_i \oplus \mathbf{z}_j \oplus (\mathbf{z}_i \odot \mathbf{z}_j) \oplus |\mathbf{z}_i - \mathbf{z}_j|\right)\right)
\label{eq:boundary}
\end{equation}

\noindent where multiplicative interaction and absolute difference capture complementary similarity patterns.

\subsubsection{Type-Specific Threshold Optimization}

The five deduplication types exhibit distinct similarity distributions. We learn type-specific thresholds through margin-based optimization:

\begin{equation}
\begin{split}
\mathcal{L}_{\text{margin}} = \sum_{t} \Bigg[ & \sum_{(i,j) \in \mathcal{P}_t^+} [m - (s_{ij} - \tau_t)]_+ \\
& + \sum_{(i,j) \in \mathcal{P}_t^-} [m - (\tau_t - s_{ij})]_+ \Bigg]
\end{split}
\label{eq:margin}
\end{equation}

\noindent where $[x]_+ = \max(0, x)$ is the hinge function and margin $m$ provides robustness against label noise.

\subsubsection{Uncertainty Quantification}

For ambiguous pairs, we quantify uncertainty through Monte Carlo dropout:

\begin{equation}
\hat{p}_{\text{dup}} = \frac{1}{M} \sum_{m=1}^{M} p_{\text{dup}}^{(m)}(d_i, d_j), \quad \sigma_{\text{dup}}^2 = \frac{1}{M} \sum_{m=1}^{M} \left(p_{\text{dup}}^{(m)} - \hat{p}_{\text{dup}}\right)^2
\label{eq:uncertainty}
\end{equation}

\noindent High variance indicates conflicting signals across feature types; these uncertain cases are routed to a more sophisticated adjudication mechanism.

\subsection{LLM-as-Judge Refinement}

\begin{figure}[htbp]
\centering
\includegraphics[width=0.5\textwidth]{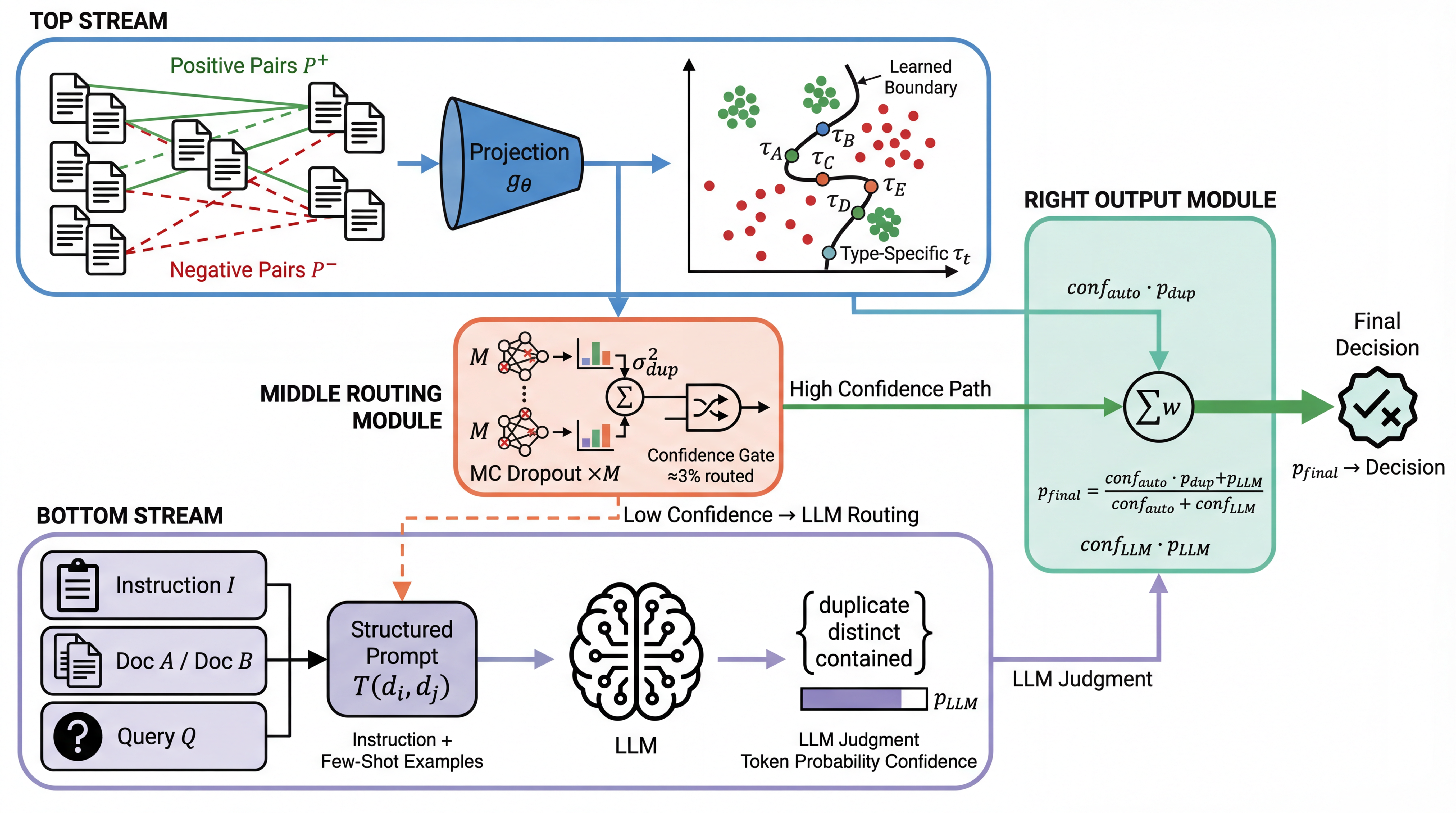}
\caption{Contrastive Boundary Learning and LLM-as-Judge refinement. The automated stream learns type-specific decision boundaries via contrastive projection and quantifies uncertainty through MC Dropout. Borderline cases ($\approx$3\% of pairs) are routed to an LLM judge, whose structured judgments are fused with automated predictions through confidence-weighted ensemble.}
\label{fig:149_4}
\end{figure}

For borderline cases where automated methods exhibit high uncertainty, we leverage the reasoning capabilities of instruction-tuned large language models. This design was motivated by the observation that humans can often resolve ambiguous duplicates through careful reading and reasoning, capabilities that LLMs increasingly approximate. The dual-stream decision architecture with uncertainty routing is shown in Fig.~\ref{fig:149_4}.

\subsubsection{Structured Prompt Engineering}

Given a candidate pair $(d_i, d_j)$, we construct a structured prompt that presents both documents along with explicit deduplication criteria:

\begin{equation}
\mathcal{T}(d_i, d_j) = [\mathcal{I}; \text{``Document A:''}, c_i; \text{``Document B:''}, c_j; \mathcal{Q}]
\label{eq:prompt}
\end{equation}

The instruction $\mathcal{I}$ encodes domain knowledge about duplicate types, and the query $\mathcal{Q}$ requests a structured judgment. We experimented with various prompt formulations and found that including concrete examples of each duplicate type significantly improved judgment accuracy, consistent with findings in few-shot prompting literature.

The LLM generates a judgment with associated probability:

\begin{equation}
p_{\text{LLM}}(y | d_i, d_j) = \text{LLM}_{\text{gen}}(\mathcal{T}(d_i, d_j))
\label{eq:llm_judge}
\end{equation}

\noindent where $y \in \{\text{duplicate}, \text{distinct}, \text{contained}\}$. We extract confidence from the token probability of the first generated token, which we found to correlate well with judgment quality.

\subsubsection{Confidence-Weighted Ensemble}

The final decision fuses automated and LLM judgments through adaptive weighting:

\begin{equation}
p_{\text{final}} = \frac{\text{conf}_{\text{auto}} \cdot p_{\text{dup}} + \text{conf}_{\text{LLM}} \cdot p_{\text{LLM}}}{\text{conf}_{\text{auto}} + \text{conf}_{\text{LLM}}}
\label{eq:ensemble}
\end{equation}

\noindent where confidence scores are derived from prediction certainty:

\begin{equation}
\text{conf}_{\text{auto}} = 1 - 2\sigma_{\text{dup}}, \quad \text{conf}_{\text{LLM}} = |p_{\text{LLM}} - 0.5| \cdot 2
\label{eq:conf}
\end{equation}

This formulation naturally trusts the more confident predictor while gracefully degrading when both are uncertain. Due to the computational cost of LLM inference, we only invoke the LLM judge when automated confidence falls below a threshold, processing approximately 3\% of candidate pairs in practice.

\subsection{Multi-Granularity Fusion Network}

Different deduplication types require different analysis granularities: Type-C duplicates exhibit character-level perturbations, Type-A template pollution requires token-level analysis, and Type-D containment is best detected at the semantic level. Our Multi-Granularity Fusion Network hierarchically combines features from all three levels.

\subsubsection{Character-Level Feature Extraction}

Character-level features capture fine-grained perturbations through multi-scale convolutions:

\begin{equation}
\mathbf{f}_{\text{char}} = \text{MaxPool}\left(\text{ReLU}\left(\sum_{k \in \mathcal{K}} \mathbf{W}_k * \mathbf{E}_{\text{char}} + \mathbf{b}_k\right)\right)
\label{eq:char}
\end{equation}

\noindent where $\mathbf{E}_{\text{char}} \in \mathbb{R}^{|c| \times d_c}$ is the character embedding matrix and $\mathcal{K} = \{3, 5, 7\}$ specifies filter widths capturing perturbations from single character substitutions to short phrase variations.

\subsubsection{Token-Level Feature Extraction}

Token-level features capture lexical and template patterns through bidirectional recurrence:

\begin{equation}
\mathbf{f}_{\text{token}} = [\overrightarrow{\mathbf{h}}_T; \overleftarrow{\mathbf{h}}_1] = \text{BiLSTM}(\mathbf{E}_{\text{token}})
\label{eq:token}
\end{equation}

\noindent The sequential inductive bias of LSTM naturally identifies template patterns at document boundaries.

\subsubsection{Semantic-Level Feature Extraction}

Semantic features project LLM embeddings through a two-layer network with GELU activation and layer normalization for training stability:

\begin{equation}
\mathbf{f}_{\text{sem}} = \text{LayerNorm}\left(\mathbf{W}_2 \cdot \text{GELU}(\mathbf{W}_1 \mathbf{e} + \mathbf{b}_1) + \mathbf{b}_2\right)
\label{eq:sem}
\end{equation}

\subsubsection{Gated Fusion Mechanism}

We employ gating to adaptively weight different granularities:

\begin{equation}
\mathbf{g} = \sigma\left(\mathbf{W}_g [\mathbf{f}_{\text{char}}; \mathbf{f}_{\text{token}}; \mathbf{f}_{\text{sem}}; \mathbf{f}_{\text{meta}}] + \mathbf{b}_g\right)
\label{eq:gate}
\end{equation}
\begin{equation}
\mathbf{f}_{\text{fused}} = \mathbf{g} \odot [\mathbf{f}_{\text{char}}; \mathbf{f}_{\text{token}}; \mathbf{f}_{\text{sem}}]
\label{eq:fused}
\end{equation}

\noindent where $\mathbf{f}_{\text{meta}}$ includes document length and vocabulary richness. Learned gate values show interpretable patterns: character-level features dominate for short documents, semantic features for long documents, and token-level features when template indicators are detected.

\subsection{Cascaded Filtering Pipeline}

\begin{figure}[htbp]
\centering
\includegraphics[width=0.5\textwidth]{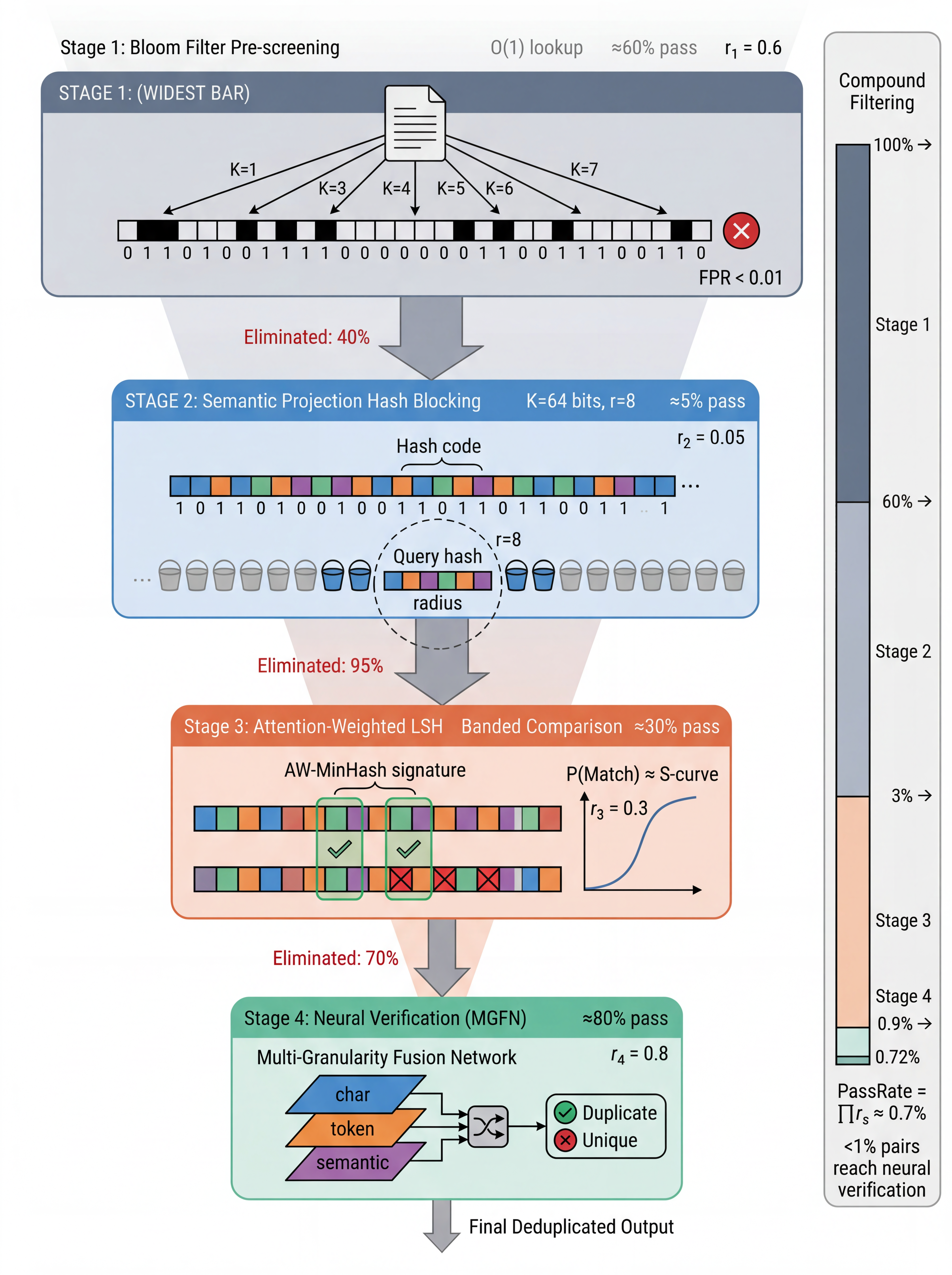}
\caption{Cascaded filtering pipeline. Four stages---Bloom filter, semantic hash blocking, attention-weighted LSH, and neural verification---reduce candidate pairs to approximately 0.7\%.}
\label{fig:149_5}
\end{figure}

Our cascaded pipeline progressively filters candidate pairs through increasingly sophisticated stages. Fig.~\ref{fig:149_5} visualizes the filtering funnel.

\subsubsection{Stage 1: Bloom Filter Pre-screening}

A Bloom filter with $k = 7$ hash functions provides $O(1)$ exact duplicate n-gram testing at $\text{FPR} < 0.01$, eliminating approximately 40\% of the corpus:

\begin{equation}
\text{FPR} = \left(1 - e^{-kn/m}\right)^k
\label{eq:bloom}
\end{equation}

\subsubsection{Stage 2: Semantic Hash Blocking}

Semantic Projection Hashing with $K = 64$ bits and Hamming radius $r = 8$ partitions documents into buckets, reducing the candidate space by approximately 95\%:

\begin{equation}
\mathcal{B}(\mathbf{z}) = \{d : \text{Hamming}(\mathbf{z}_d, \mathbf{z}) \leq r\}
\label{eq:blocking}
\end{equation}

\subsubsection{Stage 3: Attention-Weighted LSH}

AW-MinHash signatures further refine candidates through banded comparison, eliminating pairs with high Bloom filter overlap but low weighted Jaccard similarity (typically shared boilerplate):

\begin{equation}
\Pr[\text{candidate} | J_w] = 1 - \left(1 - J_w^r\right)^b
\label{eq:lsh_prob}
\end{equation}

\subsubsection{Stage 4: Neural Verification}

The Multi-Granularity Fusion Network performs final verification with decision threshold $\tau_{\text{dec}}$. The compound pass rate $\text{PassRate} = \prod_{s=1}^{4} r_s \approx 0.7\%$ (with $r_1 = 0.6$, $r_2 = 0.05$, $r_3 = 0.3$, $r_4 = 0.8$), ensuring neural computation applies to less than 1\% of all pairs.

\subsection{Containment Detection and Fragment Removal}

\subsubsection{Semantic-Enhanced Containment}

We augment lexical containment with semantic verification to avoid false positives from shared common phrases:

\begin{equation}
C_{\text{sem}}(d_i, d_j) = \frac{|G_5(d_i) \cap G_5(d_j)|}{|G_5(d_i)|} \cdot \mathbb{1}\left[\cos(\mathbf{e}_i, \mathbf{e}_j) > \tau_c\right]
\label{eq:sem_contain}
\end{equation}

\noindent Document $d_i$ is contained in $d_j$ only when both n-gram overlap exceeds 0.9 and semantic similarity confirms meaning preservation. The parent document maximizes total coverage:

\begin{equation}
d_{\text{parent}}^* = \arg\max_{d \in \mathcal{C}} \sum_{d' \in \mathcal{C} \setminus \{d\}} C_{\text{sem}}(d', d) \cdot |c_{d'}|
\label{eq:parent}
\end{equation}

\subsubsection{Viral Fragment Detection}

Type-E fragments are detected via suffix arrays over the concatenated corpus. For each high-frequency substring $f$, a semantic relevance score distinguishes meaningful content from boilerplate:

\begin{equation}
\text{Rel}(f, d) = \cos\left(f_{\text{LLM}}(f), f_{\text{LLM}}(c_d \setminus f)\right)
\label{eq:relevance}
\end{equation}

\noindent Fragments with $\text{Rel}(f, d) < \tau_{\text{rel}}$ averaged across host documents are removed; high-relevance fragments are preserved.

\section{Evaluation Metrics}

The overall performance is measured by a weighted composite score:

\begin{equation}
\text{Score}_{\text{final}} = 100 \times \sum_{t \in \{A,B,C,D,E\}} w_t \cdot S_t
\label{eq:final_score}
\end{equation}

\noindent where $w_A = 0.25$, $w_B = 0.15$, $w_C = 0.20$, $w_D = 0.20$, $w_E = 0.20$.

For near-duplicate categories (Types A, B, C), deletion recall is defined as:

\begin{equation}
S_t = \frac{\sum_{g \in \mathcal{G}_t} \text{Del}_{\text{correct}}(g)}{\sum_{g \in \mathcal{G}_t} (|g| - 1)}, \quad t \in \{A, B, C\}
\label{eq:deletion_recall}
\end{equation}

Type-D containment accuracy measures parent preservation and child removal:

\begin{equation}
S_D = \frac{\sum_{g \in \mathcal{G}_D} |\text{Children}_{\text{del}}(g)| \cdot \mathbb{1}[\text{Parent}(g) \in \mathcal{D}^*]}{\sum_{g \in \mathcal{G}_D} |\text{Children}(g)|}
\label{eq:containment_acc}
\end{equation}

Type-E fragment removal rate:

\begin{equation}
S_E = \frac{\sum_{f \in \mathcal{F}} \text{Removed}(f)}{\sum_{f \in \mathcal{F}} \text{Occurrences}(f)}
\label{eq:fragment_rate}
\end{equation}

A circuit breaker enforces Type-F preservation:

\begin{equation}
\text{Score}_{\text{final}} = 0 \quad \text{if} \quad \frac{|\mathcal{D}_F \cap \mathcal{D}^*|}{|\mathcal{D}_F|} < 0.90
\label{eq:circuit_breaker}
\end{equation}

\section{Experiment Results}

Experiments were conducted on 100GB web content from RedPajama using the MaxFrame platform. Table~\ref{tab:results} presents the comparison with baseline methods and ablation results.

\begin{table}[htbp]
\caption{Performance Comparison and Ablation Study}
\label{tab:results}
\centering
\begin{tabular}{@{}lcccccc@{}}
\toprule
Method & $S_A$ & $S_B$ & $S_C$ & $S_D$ & $S_E$ & Final \\
\midrule
SimHash-64 & 0.71 & 0.65 & 0.42 & 0.31 & 0.18 & 48.20 \\
MinHash-LSH & 0.82 & 0.78 & 0.68 & 0.45 & 0.22 & 61.35 \\
NearDup-BERT & 0.85 & 0.79 & 0.74 & 0.68 & 0.55 & 73.45 \\
DedupLM & 0.88 & 0.83 & 0.81 & 0.76 & 0.71 & 81.20 \\
\midrule
SemHash-LLM & 0.94 & 0.91 & 0.89 & 0.92 & 0.88 & 91.05 \\
\quad w/o SPH & 0.83 & 0.80 & 0.79 & 0.74 & 0.70 & 79.32 \\
\quad w/o AW-MinHash & 0.89 & 0.85 & 0.82 & 0.88 & 0.83 & 85.18 \\
\quad w/o MGFN & 0.86 & 0.82 & 0.80 & 0.85 & 0.79 & 84.56 \\
\bottomrule
\end{tabular}
\end{table}

\section{Conclusion}

This paper presented SemHash-LLM, a semantic-aware deduplication framework integrating large language model capabilities with efficient hashing techniques. The framework achieves state-of-the-art performance across all five deduplication categories while maintaining computational efficiency at trillion-scale.
\bibliographystyle{IEEEtran}
\bibliography{references}
\end{document}